\documentclass{article} % For LaTeX2e
\usepackage{iclr2024_conference,times}
\iclrfinalcopy

\usepackage{amsmath,amsfonts,bm}

\def\eqref#1{equation~\ref{#1}}
\def\1{\bm{1}}

\DeclareMathAlphabet{\mathsfit}{\encodingdefault}{\sfdefault}{m}{sl}
\SetMathAlphabet{\mathsfit}{bold}{\encodingdefault}{\sfdefault}{bx}{n}

\usepackage[utf8]{inputenc} % allow utf-8 input
\usepackage[T1]{fontenc}    % use 8-bit T1 fonts
\usepackage{hyperref}       % hyperlinks
\usepackage{url}            % simple URL typesetting
\usepackage{booktabs}       % professional-quality tables
\usepackage{amsfonts}       % blackboard math symbols
\usepackage{nicefrac}       % compact symbols for 1/2, etc.
\usepackage{microtype}      % microtypography
\usepackage{xcolor}         % colors
\usepackage{booktabs}

\usepackage{multirow}
\usepackage{arydshln}
\usepackage{verbatim}
\usepackage{sourcecodepro}
\usepackage{tcolorbox}
\tcbuselibrary{raster}
\tcbuselibrary{breakable}
\usepackage{graphicx}
\usepackage[framemethod=TikZ]{mdframed}
\usepackage{tikz}
\usepackage{tikzsymbols}
\usepackage{adjustbox}
\usetikzlibrary{arrows.meta}
\usetikzlibrary{shapes.misc, positioning}
\usepackage{pgfplots}
\usepackage{caption}
\usepackage{subcaption}
\usepackage{cleveref}
\usepackage{listings}
\usepackage{soul}

\title{Universal Self-Consistency for Large Language Model Generation}

\author{
Xinyun Chen~\thanks{Equal contribution.} \hspace{.5em}
Renat Aksitov \textsuperscript{*} \hspace{.5em}
Uri Alon \hspace{.5em}
Jie Ren \hspace{.5em}
Kefan Xiao \hspace{.5em}
Pengcheng Yin \hspace{.5em} \\
\textbf{
Sushant Prakash \hspace{.5em}
Charles Sutton \hspace{.5em}
Xuezhi Wang \hspace{.5em}
Denny Zhou} \\ \\ [0.5ex]
\hspace{60mm} Google
}

\renewcommand{\headrulewidth}{1pt}
\fancypagestyle{firstpage}{
  \lhead{\begin{picture}(0,0)\put(0,-3){\includegraphics[width=0.25\linewidth]{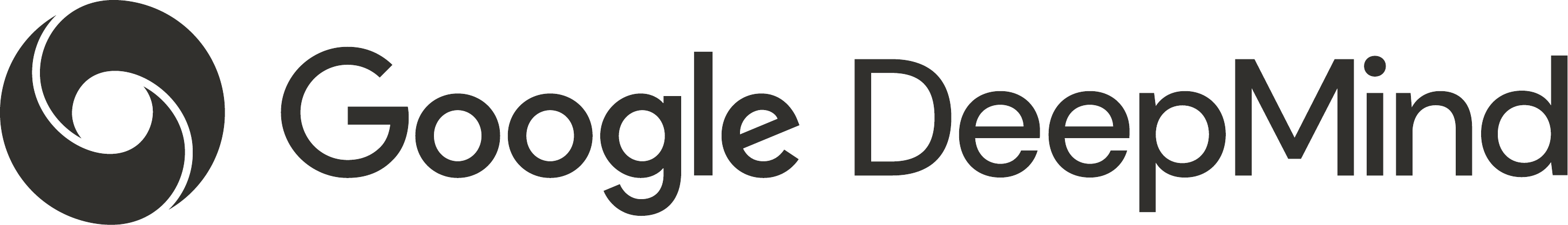}}\end{picture}}
}

\begin{document}

\maketitle

\newcommand{\fix}{\marginpar{FIX}}
\newcommand{\new}{\marginpar{NEW}}
\newcommand{\eat}[1]{}

\definecolor{medblue}{rgb}{0,0,.75}
\definecolor{burntorange}{rgb}{0.8, 0.33, 0.0}

\renewcommand*{\ttdefault}{qcr}

\newif\ifsubmit
\submitfalse

\ifsubmit
\newcommand{\xinyun}[1]{}
\newcommand{\uri}[1]{}
\newcommand{\jie}[1]{}
\newcommand{\urix}[1]{}
\newcommand{\xuezhi}[1]{}
\newcommand{\renat}[1]{}
\else
\newcommand{\xinyun}[1]{\textcolor{blue}{[Xinyun: #1]}}
\newcommand{\uri}[1]{\textcolor{magenta}{[Uri: #1]}}
\newcommand{\jie}[1]{\textcolor{magenta}{[Jie: #1]}}
\newcommand{\urix}[1]{\textcolor{magenta}{\st{#1}}}
\newcommand{\xuezhi}[1]{\textcolor{red}{[Xuezhi: #1]}}
\newcommand{\renat}[1]{\textcolor{purple}{[Renat: #1]}}
\fi

\thispagestyle{firstpage}

\begin{abstract}
Self-consistency with chain-of-thought prompting (CoT) has demonstrated remarkable performance gains on various challenging tasks, by utilizing multiple reasoning paths sampled from large language models (LLMs). However, self-consistency relies on the answer extraction process to aggregate multiple solutions, which is not applicable to free-form answers. In this work, we propose Universal Self-Consistency (USC), which leverages LLMs themselves to select the most consistent answer among multiple candidates. We evaluate USC on a variety of benchmarks, including mathematical reasoning, code generation, long-context summarization, and open-ended question answering. On open-ended generation tasks where the original self-consistency method is not applicable, USC effectively utilizes multiple samples and improves the performance. For mathematical reasoning, USC matches the standard self-consistency performance without requiring the answer formats to be similar. Finally, without access to execution results, USC also matches the execution-based voting performance on code generation.
%\uri{I'd switch the order of the last two contributions: improving on open-ended tasks sounds like a more novel achievement than only "matching" existing results}
\end{abstract}

\section{Introduction}
\label{sec:intro}

Large language models (LLMs) have accomplished significant breakthroughs in a wide variety of domains, including mathematical reasoning~\citep{cobbe2021training, wei2022chain,lewkowycz2022solving}, code generation~\citep{chen2021evaluating,austin2021program,li2022competition}, and other text generation tasks~\citep{bubeck2023sparks,anil2023palm,touvron2023llama}. Despite the rapid progress, the LLM-generated responses are still prone to errors when they get long. A long line of efforts have been devoted to improve the output quality by sampling multiple model responses and then selecting the final output based on certain criteria. For example, prior works have trained neural networks to rerank model outputs~\citep{cobbe2021training,li2023making,ni2023lever,yin2019reranking,zeng2022n}, and more recent works investigate using LLMs to score the responses~\citep{fu2023gptscore,liu2023geval,wang2023chatgpt}.

In this work, we consider the \emph{consistency} among model responses as the criterion to select the model output, a generic metric that has enabled huge performance leaps in reasoning~\citep{wang2022self} and code generation~\citep{li2022competition,shi-etal-2022-natural}. In particular, self-consistency \citep{wang2022self} with chain-of-thought prompting~\citep{wei2022chain} boosts the performance on various  benchmarks, by marginalizing latent reasoning paths through sampling which leads to select the final answer as the most common one.
However, self-consistency can only be applied to tasks where the final answer can be aggregated via exact match, e.g., a single number for math problems.

To address this major limitation of self-consistency,  we propose Universal Self-Consistency (USC) to support various applications, especially free-form generation tasks. Specifically, given multiple candidate responses, USC simply calls the LLM to select the most consistent response among them as the final output. Thus, USC eliminates the need of designing an answer extraction process, and is applicable to tasks with free-form answers. Although prior works have revealed weaknesses of LLMs for response selection, such as position bias~\citep{wang2023large,zheng2023large} and incorrectly judging the answer correctness~\citep{huang2023large,gou2023critic}, intuitively, assessing the consistency among candidate answers is easier than measuring and comparing the answer quality.

We evaluate universal self-consistency on a wide range of tasks, including mathematical reasoning, code generation, long-context summarization, and open-ended question answering. On GSM8K~\citep{cobbe2021training} and MATH~\citep{hendrycks2021measuring} benchmarks for math problem solving, USC generally matches the performance of the standard self-consistency. On programming tasks including text-to-SQL generation~\citep{li2023can} and Python code generation~\citep{yin-etal-2023-natural}, USC matches the performance of execution-based consistency~\citep{li2022competition,shi-etal-2022-natural}, while USC does not require execution results to aggregate over candidate programs. Finally, USC also improves the performance for open-ended question answering~\citep{lin2021truthfulqa} and long-context summarization~\citep{huang2021efficient,chen2022summscreen}, where the standard self-consistency is not applicable. In addition to the performance gain, our evaluation also demonstrates that USC outputs highly match those of the standard self-consistency when the comparison is applicable, while it is robust to the ordering of candidate responses.
\section{Background: Self-Consistency}
\label{sec:background}

Self-consistency \citep{wang2022self} augments chain-of-thought prompting \citep{wei2022chain} by sampling multiple reasoning chains and then taking a majority vote on the final answer set.
The intuition is that sometimes the greedily decoded reasoning process might not be the optimal one, hence it makes more sense to sample a diverse set of reasoning chains, and if some of them lead to the same answer, then we have a higher confidence that this consistent answer is the correct one.
It has been shown that self-consistency improves the greedy  chain-of-thought prompting by a large margin on a wide set of reasoning tasks.

Besides question answering tasks, consistency-based answer selection has also been applied to code generation~\citep{shi-etal-2022-natural,li2022competition,chen2019execution}, which requires code execution. Specifically, we first execute all predicted programs on the given inputs, then programs with the same execution outputs are clustered together, assuming that they are semantically equivalent. Finally, we select the program belonging to the largest cluster as the final prediction. When the program inputs given in the task description are insufficient to distinguish between different predictions, this execution-based code selection is also often accompanied with a test case generation process to better examine the consistency~\citep{li2022competition,chen2022codet,huang2023enhancing}. 

% \urix{\paragraph{Limitations.} }
Despite the remarkable improvement, self-consistency is only applicable to problems with a unique and closed-form answer, e.g., when the final answer consists of a single number, because a majority vote needs to be taken over the final answer set. 
% In addition, the answer extraction process in self-consistency requires all samples to share a similar format that represents the final answer.
This significant requirement poses a challenge for tasks that require open-ended generations, such as summarization, creative writing, and open-ended question answering.
% \uri{I don't understand the problem of answer extraction and having a similar format. In both SC and USC, we need to prompt/instruct the model to present the answer in a certain format ("The answer is: " or "The chosen summary is: "), so what's the difference?} \renat{No, with USC having the same format is not required (other than to be able to do parallel evaluation with SC).}
% \uri{You're saying that you get 8 initial outputs, one of them says "the \textbf{answer} is: 8" and the other says "the \textbf{result} is: 8", and you can just feed all of them as-is for USC? OK, but doesn't sound to me like an high-level advantage} \renat{I'm thinking more about something like MATH, where parsing is extremely challenging if the model wasn't trained to output answer in the predefined formats. Here is a parser that supports GPT-4 outputs on MATH - https://github.com/chuanyang-Zheng/Progressive-Hint/blob/main/utils.py}
% \uri{OK, but it should be understandable to the reader and sound important without reading external code}
\section{Universal Self-Consistency}
\label{sec:approach}

\begin{figure}[h]
\centering
  \includegraphics[width=\linewidth]{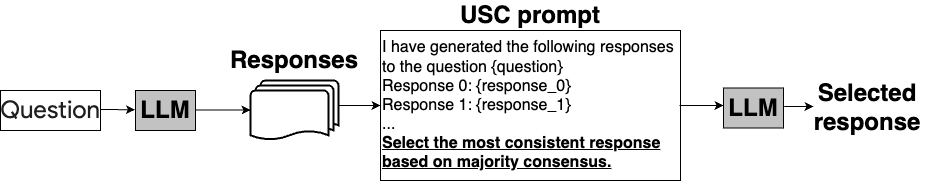}
  \caption{Overview of the Universal Self-Consistency workflow.}
\label{fig:usc}
\end{figure}

\begin{figure}[h]
\centering
\begin{subfigure}[b]{\textwidth}
  \includegraphics[width=\linewidth]{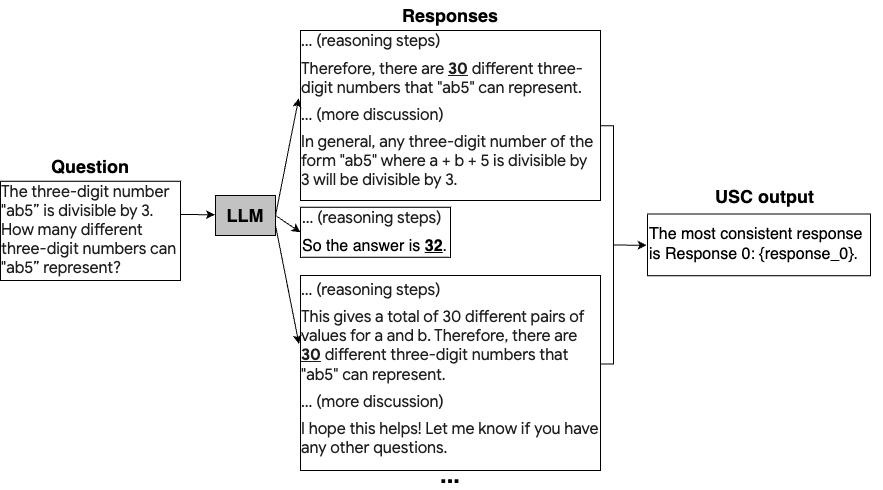}
  \caption{}
  \label{fig:usc-math}
\end{subfigure}
\begin{subfigure}[b]{\textwidth}
  \includegraphics[width=\linewidth]{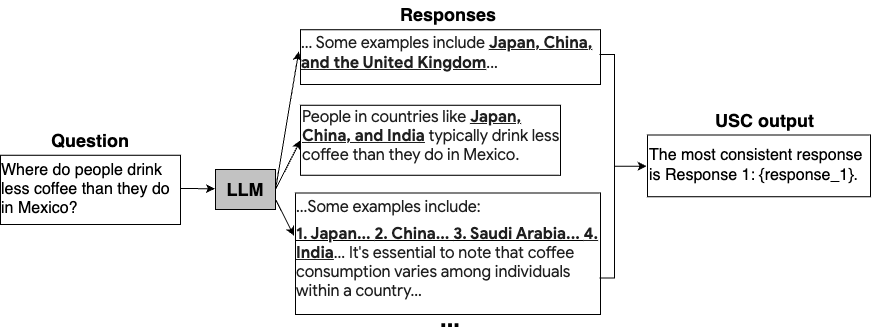}
  \caption{}
  \label{fig:usc-qa}
\end{subfigure}
  \caption{Examples of Universal Self-Consistency for answer selection from responses of diverse formats: (a) mathematical reasoning; and (b) open-ended question answering. Note that for the given open-ended question, the final answer is an entity list, where no two responses share the same predictions. Still, the LLM correctly selects the response where the individual entities in the predicted list appear most frequently in the candidate responses.}
\label{fig:usc-tasks}
\end{figure}

We present the overall workflow of universal self-consistency (USC) in Figure~\ref{fig:usc}, which utilizes LLMs to enable self-consistency for a wide variety of tasks, especially free-form text generation.
First, we sample multiple responses with the large language model. Afterward, to select one model response as the final answer, we concatenate all responses together, and then construct a prompt with an instruction asking the language model to select the most consistent response. In this way, USC obviates the necessity of counting the exact answer frequency as in the standard self-consistency, and relies on the LLM's own ability to measure the consistency among different responses. Although prior works show that LLMs sometimes have trouble evaluating the prediction correctness~\citep{huang2023large,gou2023critic}, especially for reasoning problems, empirically we observe that LLMs are generally able to examine the response consistency across multiple tasks.

Consistency assessment with LLMs offers more flexibility for free-form generation. Figure~\ref{fig:usc-tasks} demonstrates example tasks where different consistency criteria are beneficial for response selection. Specifically, Figure~\ref{fig:usc-math} shows different model responses for a math problem, where the output formats are diverse and thus makes it challenging for rule-based methods to extract answers. Nonetheless, assuming that the final answers are correctly extracted, the consistency criterion still follows the standard self-consistency on mathematical reasoning, which is based on the exact match of the final answers represented as single numerical values. On the other hand, Figure~\ref{fig:usc-qa} shows an example question where the final answer is an entity list. Despite that there is no response that is consistent with others based on the exact match, the LLM selects the response where each of the predicted entities appears most frequently among the candidate outputs. In Section~\ref{sec:exp}, we further show that LLM can also examine the consistency among responses beyond the question answering tasks, including code generation without access to the execution outputs, and long-context summarization.
\section{Experiments}
\label{sec:exp}

\subsection{Evaluation Setup}

\paragraph{Benchmarks.} We evaluate USC on the following variety of tasks:
\begin{itemize}
\item \emph{Mathematical reasoning benchmarks}, including GSM8K~\citep{cobbe2021training}, a dataset of 8,500 grade school math word problems, and MATH~\citep{hendrycks2021measuring}, a dataset of 12,500 challenging mathematics problems from high school competitions.
\item \emph{Code generation benchmarks}, including BIRD-SQL dataset~\citep{li2023can} for text-to-SQL generation, and ARCADE dataset~\citep{yin-etal-2023-natural} for Python code generation in data science notebooks.
\item \emph{Long-context summarization}, including the GovReport and SummScreen  benchmarks from ZeroSCROLLS \citep{shaham2023zeroscrolls}. 
In GovReport \citep{huang2021efficient}, each input is a document containing $\sim$7,900 words on average, and the reference output is an expert-written executive summary with $\sim$500 words. 
In SummScreen \citep{chen2022summscreen}, every input is a transcript of a TV show episode with $\sim$5,600 words, and each reference output is a $\sim$100 words human-written recap of the episode.
We follow \citet{shaham2023zeroscrolls} and measure ROUGE 1, ROUGE 2, and ROUGE-Lsum which measure n-gram overlap with the reference summary, and we also measure BERTScore F1 \citep{zhang2019bertscore}.
\item \emph{TruthfulQA \citep{lin2021truthfulqa} benchmark} for open-ended question answering, which contains 817 questions to test model's ability in generating truthful answers.
To evaluate the answer's quality, we use the GPT-judge and GPT-info, which are GPT-3 models fine-tuned on human feedback data, provided by \cite{lin2021truthfulqa}.
GPT-judge model outputs a binary rating for truthfulness, and GPT-info model outpus a binary rating for informativeness.
It is shown that the GPT-3 models have higher accuracy in predicting human judgement than the automatic metrics ROUGE, BLEU, BLEURT.
\end{itemize}

\paragraph{Decoding schemes.} We compare USC to the following decoding schemes:

\begin{itemize}
    \item \emph{Greedy decoding} generates a single answer with the temperature 0.
    \item \emph{Random} selects one answer randomly from multiple samples with temperature > 0.
    %\uri{That's technically how we did it, but conceptually isn't it equivalent to just "sampling one output with a non-zero temperature"?}~\xinyun{I was trying to note that the sample set is the same for all sampling method, so I describe the method in this way.} \uri{ok}
    \item \emph{SC}~\citep{wang2022self} is the standard self-consistency decoding with answer extraction. We evaluate SC whenever applicable; for example, on reasoning benchmarks where the final answers can be compared through exact match.
\end{itemize}

To enable a fair comparison, for sampling schemes (i.e., except greedy decoding), we always select the final answer from the same set of initial model responses. For code generation, we compare our approach to execution-based self-consistency~\citep{shi-etal-2022-natural,li2022competition,chen2019execution}, where we select the code with the most common execution result. Both USC and execution-based self-consistency first filter out syntactically invalid candidate programs, and then perform the voting over the remaining ones.  For ARCADE benchmark, we also evaluate a variant of the execution-based self-consistency with fuzzy matching as described in~\citet{yin-etal-2023-natural}, which implements a set of heuristics to determine whether the execution outputs of two programs are equivalent when they are not exact match.

\eat{
\uri{\paragraph{Baselines}
Across all tasks, we compare to ``greedy'' decoding, which uses the same base model with a sampling temperature of zero. Another straightforward baseline is ``one sample'', which samples one output with the same (larger than zero) temperature we used to generate our initial outputs;
and we also compare to ``top sample'', which samples the same number of initial outputs as USC, but chooses the one with the highest log-probability. \xinyun{For the initial submission, I suggest that we exclude this comparison. It is only applicable to Google models (probably we can also evaluate open-source LLMs later), and it makes the author institution more explicit.}
Whenever possible, we compare to the standard Self-Consistency \citep{wang2022self}; however in open-ended tasks such as summarization, standard Self-Consistency \emph{cannot be applied} because no outputs are compltely identical.
}}

\paragraph{Implementation details.} We conduct experiments using instruction-tuned \texttt{PaLM 2-L}~\citep{anil2023palm} and \texttt{gpt-3.5-turbo} models. Unless otherwise specified, the LLM generates 8 initial samples for both SC and USC. For mathematical reasoning, summarization and the ARCADE benchmark for Python code generation, the initial samples are generated with zero-shot prompting, thus the output formats are diverse. For BIRD-SQL, we used the 1-shot chain-of-thought prompt in~\citet{li2023can}, which improves the performance. We also utilized a one-shot prompt for TruthfulQA to improve the quality of candidate responses.
We set the temperature to be 0.6 for \texttt{PaLM 2-L}, and 1.0 for \texttt{gpt-3.5-turbo}.

\eat{\renat{We use t=0.6 for PaLM and t=1.0 for gpt-3.5. GSM8K and MATH are completely zero shot, only problem or question are given as an input. MMLU is zero shot with instructions, to set up multi-choice task}}

% \eat{\uri{I'm not sure that in the outside world people know what "2-L" mean (and it also kind of breaks affiliation anonymity?). Can we call it "PaLM-2 340B"? I don't think it's classified info.}
% \xuezhi{The largest PaLM model (340b) is called ``PaLM 2-L'' in the official PaLM 2 technical report. We didn't disclose the model size (340b) in the report so we probably shouldn't say 340b anywhere in any of our papers.} \uri{agree, thanks}}

\subsection{Main Results}

\paragraph{Mathematical reasoning.}
For mathematical reasoning benchmarks, we compare USC against the standard self-consistency in \Cref{tab:result-reasoning}. For the standard self-consistency, we employ a regular expression matching to extract the final answer on GSM8K, and re-use the answer parsing code from \citep{zheng2023progressive} for MATH. Overall, USC consistently improves over the greedy decoding and random selection, and the performance is generally comparable to the standard self-consistency, which USC does not need answer parsing to perform the voting.

\paragraph{Code generation.} Table~\ref{tab:result-code} presents the results on BIRD-SQL and ARCADE respectively. On BIRD-SQL, besides the execution accuracy, we follow~\citep{li2023can} to also evaluate the valid efficiency score, which measures the efficiency of the generated SQL queries. We show that USC matches the execution-based self-consistency performance on both benchmarks, while USC does not utilize code execution to perform the voting.

\paragraph{Summarization.}
Results for the summarization benchmarks are shown in \Cref{tab:result-summarization}.
Since the generated summaries are in free-form, the standard self-consistency is not applicable. In GovReport, USC consistently improves over the baselines across all metrics. In Section~\ref{sec:ablations}, we further show that asking the model to choose the \emph{most detailed} summary results in more performance gain.

\paragraph{TruthfulQA.}
Table \ref{tab:result-truthfulqa} presents results on TruthfulQA, where SC is also not directly applicable because the generated answers are in free-form. Comparing with greedy decoding and random selection, USC-based answers have the highest truthfulness with both \texttt{PaLM 2-L} and \texttt{gpt-3.5-turbo}. For informativeness which is considered as a secondary objective, USC-based answers have the highest score on \texttt{PaLM 2-L} and the second highest score (0.1 lower than the highest) on \texttt{gpt-3.5-turbo}.
Considering that GPT-judge and GPT-info models have generally ~90-95\% validation accuracy on rating prediction \citep{lin2021truthfulqa}, the 0.1 difference is not considered significant.
\begin{table}[h]
\centering
\caption{Accuracy on mathematical reasoning benchmarks. USC and SC consistently improve over the greedy decoding and random selection. USC performance is generally comparable to SC.}
\label{tab:result-reasoning}
\begin{tabular}{clcc}
\toprule
Model & Approach & GSM8K & MATH \\
\midrule
\multirow{4}{*}{\texttt{PaLM 2-L}} & Greedy decoding & 85.7 & 30.8 \\
& Random & 82.9 & 28.0 \\
& SC~\citep{wang2022self} & \textbf{90.4} & \textbf{37.9} \\
& USC & 90.2 & 37.4 \\
\midrule
\multirow{4}{*}{\texttt{gpt-3.5-turbo}} & Greedy decoding & 73.4 & 33.2 \\
& Random & 68.5 & 26.3 \\
& SC & \textbf{78.5} & 38.0\\
& USC & 77.8 & \textbf{38.1}\\
\bottomrule
\end{tabular}
\end{table}

\begin{table}[h]
\centering
\caption{Accuracy on code generation benchmarks with \texttt{gpt-3.5-turbo}.}
\label{tab:result-code}
\begin{tabular}{clcc}
\toprule
Dataset & Approach & Execution Accuracy & Valid Efficiency Score \\
\midrule
\multirow{4}{*}{BIRD-SQL} & Greedy decoding & 42.4 & 44.4 \\
& Random & 41.9 & 44.0 \\
& SC-Exec & \textbf{45.6} & 48.1 \\
& USC & 45.5 & \textbf{48.8}\\
\midrule
\multirow{5}{*}{ARCADE} & Greedy decoding & 26.0 & \multirow{5}{*}{N/A} \\
& Random & 26.8\\
& SC-Exec (strict match) & 29.8\\
& SC-Exec (fuzzy match) & \textbf{30.3}\\
& USC & 30.1\\
\bottomrule
\end{tabular}
\end{table}

\begin{table}[h]
\centering
\caption{Results on long-context summarization benchmarks with \texttt{PaLM 2-L}. Since the outputs are in free-form, the standard self-consistency is not applicable. USC consistently improves over the baselines on summary quality.}
\label{tab:result-summarization}
\begin{tabular}{clcccc}
\toprule
Dataset & Approach & ROUGE-1 & ROUGE-2 & ROUGE-Lsum & BERTScore \\
\midrule
\multirow{3}{*}{GovReport} & Greedy decoding & 38.8 & 16.9 & 33.8 & 62.7  \\
 & Random & 38.5 & 16.9 & 33.6 & 62.6 \\
 & USC & \textbf{40.2} & \textbf{17.4} & \textbf{35.1} & \textbf{62.8} \\
\midrule
\multirow{3}{*}{SummScreen} & Greedy decoding & 30.6 & 7.5 & 19.1 & \textbf{58.7}  \\
& Random & 30.2 & 7.3 & 19.0 & 58.6  \\
& USC & \textbf{31.7} & \textbf{7.8} & \textbf{19.8} & 58.3 \\
\bottomrule
\end{tabular}
\end{table}

\begin{table}[h]
\centering
\caption{Accuracy on the TruthfulQA benchmark. Since the answer is in free-form, the standard self-consistency is not applicable. USC overall has the highest truthfulness and informativeness over the baselines.}
\label{tab:result-truthfulqa}
\begin{tabular}{clcc}
\toprule
Model & Approach & GPT-judge & GPT-info \\
\midrule
\multirow{3}{*}{\texttt{PaLM 2-L}} & Greedy decoding & 62.1 & 95.1 \\
& Random & 62.9 & 94.6 \\
& USC & \textbf{67.7} & \textbf{99.0} \\
\midrule
\multirow{3}{*}{\texttt{gpt-3.5-turbo}} & Greedy decoding & 79.8 & \textbf{99.7} \\
& Random & 80.6 & 99.3 \\
& USC & \textbf{82.5} & 99.6\\
\bottomrule
\end{tabular}
\end{table}

\subsection{Ablations}
\label{sec:ablations}

\paragraph{Effect of response ordering.} Prior works have shown that large language models can be affected by the order of candidate responses when used to evaluate their quality~\citep{wang2023large,zheng2023large}. We examine the effect of response ordering by performing USC with 5 different random orders when concatenating all responses, and calculate the mean and standard deviation of the task results. From Table~\ref{tab:result-reasoning-variance}, we observe that the overall model performance remains similar with different response orders, suggesting the effect of response order is minimal. 

\begin{table}[h!]
\caption{USC performance with random shuffling of original responses using \texttt{PaLM 2-L}. The mean and standard deviation are computed with 5 runs.}
\label{tab:result-reasoning-variance}
\small
\begin{subfigure}[t]{.2\textwidth}
\centering
\caption{}
\label{tab:result-reasoning-variance-a}
% \begin{minipage}{1\textwidth}
\begin{tabular}{lc}
\toprule
Dataset & Acc \\
\midrule
GSM8K & 89.7\scriptsize{$\pm$0.3}  \\
MATH & 37.3\scriptsize{$\pm$0.2}  \\
\bottomrule
\end{tabular}
% \end{minipage}
\end{subfigure}
\hfill
\begin{subfigure}[t]{.4\textwidth}
\centering
\caption{}
\label{tab:result-reasoning-variance-b}
% \begin{minipage}{1\textwidth}
\begin{tabular}{lcc}
\toprule
Dataset & ROUGE-1 & ROUGE-Lsum \\
\midrule
SummScreen & 31.6\scriptsize{$\pm$0.3} & 19.5\scriptsize{$\pm$0.2} \\
GovReport & 40.0\scriptsize{$\pm$0.1} & 34.9\scriptsize{$\pm$0.2} \\
\bottomrule
\end{tabular}
% \end{minipage}
\end{subfigure}
\hfill
\begin{subfigure}[t]{0.3\textwidth}
\centering
% \begin{minipage}{1\textwidth}
\caption{}
\label{tab:result-reasoning-variance-c}
\begin{tabular}{lc}
\toprule
Metric & TruthfulQA \\
\midrule
GPT-judge & 68.3\scriptsize{$\pm$0.6}  \\
GPT-info & 99.0\scriptsize{$\pm$0.1}  \\
\bottomrule
\end{tabular}
% \end{minipage}
\end{subfigure}
\end{table}

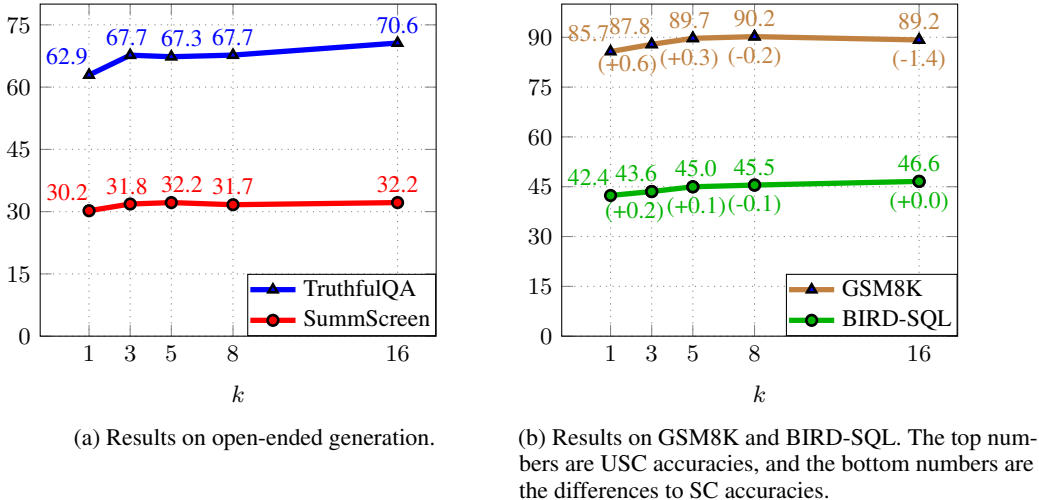
\begin{figure}[h!]
\begin{subfigure}[t]{0.49\textwidth}
    \begin{tikzpicture}[scale=1]
      \begin{axis}[
        xlabel={$k$},
        xlabel style={font=\footnotesize},
        ylabel style={font=\footnotesize},
        yticklabel style={font=\footnotesize}, 
        xticklabel style={font=\footnotesize}, 
        ylabel near ticks,
        legend style={at={(1,0.)},anchor=south east,mark size=2pt,font=\footnotesize,
        inner xsep=1pt, inner sep=0pt},
            legend cell align={left},
        legend cell align={left},
        xmin=-0.5, xmax=17,
        ymin=0, ymax=80,
        xtick={1,3,5,8,16},
        ytick={0,15,...,100},
          % ylabel shift={-5pt},        
          grid = major,
          major grid style={dotted,gray},
          height=6cm,
          width=1\textwidth,
          enlarge x limits=0.05,
          ]
          
            \addplot[color=blue, solid, mark options={solid, fill=blue, draw=black, line width=1pt}, mark=triangle*, line width=2pt, mark size=2pt, visualization depends on=\thisrow{alignment} \as \alignment, nodes near coords, point meta=explicit symbolic,
            every node near coord/.style={anchor=\alignment, font=\footnotesize}] 
            table [meta index=2]  {
            x   y       label   alignment
1   62.91   {62.9}  -40
3   67.69   {67.7}  -80
5   67.32   {67.3}  -120
8   67.71   {67.7}  -90
16  70.62   {70.6}  -90
  };
  \addlegendentry{TruthfulQA}
  
    \addplot[color=red, solid, mark options={solid, fill=red, draw=black, line width=1pt}, mark=*, line width=2pt, mark size=2pt, visualization depends on=\thisrow{alignment} \as \alignment, nodes near coords, point meta=explicit symbolic,
          every node near coord/.style={anchor=\alignment, font=\footnotesize}] 
          table [meta index=2]  {
            x   y       label   alignment
1   30.22   {30.2}  -40
3   31.84   {31.8}  -80
5   32.18   {32.2}  -120
8   31.67   {31.7}  -90
16  32.18   {32.2}  -90
            }; 
            \addlegendentry{SummScreen}

          \end{axis}
  \end{tikzpicture}
 \caption{Results on open-ended generation.}
  \label{fig:sampling-gen} 
\end{subfigure}
\begin{subfigure}[t]{0.49\textwidth}
    \begin{tikzpicture}[scale=1]
      \begin{axis}[
        xlabel={$k$},
        xlabel style={font=\footnotesize},
        ylabel style={font=\footnotesize},
        yticklabel style={font=\footnotesize}, 
        xticklabel style={font=\footnotesize}, 
        ylabel near ticks,
        legend style={at={(1,0.)},anchor=south east,mark size=2pt,font=\footnotesize,
        inner xsep=1pt, inner sep=0pt},
            legend cell align={left},
        legend cell align={left},
        xmin=-0.5, xmax=17,
        ymin=0, ymax=100,
        xtick={1,3,5,8,16},
        ytick={0,15,...,100},
          % ylabel shift={-5pt},        
          grid = major,
          major grid style={dotted,gray},
          height=6cm,
          width=1\textwidth,
          enlarge x limits=0.05,
          ]

            \addplot[color=brown, solid, mark options={solid, fill=blue, draw=black, line width=1pt}, mark=triangle*, line width=2pt, mark size=2pt, visualization depends on=\thisrow{alignment} \as \alignment, nodes near coords, point meta=explicit symbolic,
            every node near coord/.style={anchor=\alignment, font=\footnotesize}] 
            table [meta index=2]  {
            x   y       label   alignment
1   85.7    {85.7}  -40
3   87.87   {(+0.6)}  40
3   87.87   {87.8}  -40
5   89.69   {(+0.3)}  80
5   89.69   {89.7}  -80
8   90.2   {(-0.2)}  90
8   90.2   {90.2}  -90
16  89.2   {(-1.4)}  90
16  89.2   {89.2}  -90
  };
  \addlegendentry{GSM8K}          

            \addplot[color=green!70!black, solid, mark options={solid, fill=green!70!black, draw=black, line width=1pt}, mark=*, line width=2pt, mark size=2pt, visualization depends on=\thisrow{alignment} \as \alignment, nodes near coords, point meta=explicit symbolic,
          every node near coord/.style={anchor=\alignment, font=\footnotesize}] 
          table [meta index=2]  {
            x   y       label   alignment
1   42.37   {42.4}  -40
3   43.55   {(+0.2)}  50
3   43.55   {43.6}  -50
5   44.98   {(+0.1)}  100
5   44.98   {45.0}  -100
8   45.50   {(-0.1)}  90
8   45.50   {45.5}  -90
16  46.61   {(+0.0)}  90
16  46.61   {46.6}  -90
            }; 
            \addlegendentry{BIRD-SQL}

          \end{axis}

  \end{tikzpicture}
 \caption{Results on GSM8K and BIRD-SQL. The top numbers are USC accuracies, and the bottom numbers are the differences to SC accuracies.}
  \label{fig:sampling-w-sc} 
\end{subfigure}
 \caption{USC results with different number of samples.}
  \label{fig:sampling} 
\end{figure}

\paragraph{Different number of responses.} Next, we examine the effect of using different numbers of responses in USC. As shown in \Cref{fig:sampling}, USC consistently benefits from more samples on TruthFulQA and BIRD-SQL. However, USC does not further improve the performance on SummScreen after 5 samples, and the accuracy on GSM8K decreases with 16 samples.
This can be due to the weakness in long-context understanding when the prompt contains more candidate responses, and the imperfect counting ability of LLMs. Nevertheless, we consider utilizing a few samples (e.g., 8) a sweet spot to balance the task accuracy and compute cost, in which case USC reliably improves the performance across the board. In Section~\ref{sec:usc-vs-sc}, we further compare the predictions from USC and SC to understand how using more candidate responses affects the results.

\paragraph{Criteria for response selection.} 
    One advantage of USC is its generality: the same criteria can be applied to various tasks, without any task-specific knowledge. Nonetheless, a minor task-specific adaptation of the response selection instruction can further boost USC over the generic prompts. For example, \Cref{tab:result-summarization-detailed} shows that asking the LLM to choose the most \emph{detailed} response (rather than the most \emph{consistent} one) results in gains of about 2 ROUGE-1 and ROUGE-Lsum points.

\begin{table}[h]
\centering
\caption{Ablation on the response selection criterion on long-context summarization benchmarks with \texttt{PaLM 2-L}.}
\label{tab:result-summarization-detailed}
\begin{tabular}{clcccc}
\toprule
Dataset & Approach & ROUGE-1 & ROUGE-2 & ROUGE-Lsum & BERTScore \\
\midrule
\multirow{2}{*}{GovReport}  & USC & 40.2 & 17.4 & 35.1 & 62.8 \\
& USC -- most detailed & \textbf{42.4} & \textbf{18.2} & \textbf{36.9} & \textbf{63.2}\\
\midrule
\multirow{2}{*}{SummScreen} & USC & 31.7 & 7.8 & 19.8 & \textbf{58.3} \\
 & USC -- most detailed & \textbf{33.0} & \textbf{7.9} & \textbf{22.0} & \textbf{58.3} \\
\bottomrule
\end{tabular}
\end{table}

\subsection{Discussion: How well does USC match SC selection?}
\label{sec:usc-vs-sc}

% \begin{table}[h]
% \centering
% \caption{Comparison of selections made by USC versus SC with PaLM 2-L.}
% \label{tab:usc_vs_sc}
% \begin{tabular}{c p{1cm} c p{1.8cm} p{1.8cm} p{1.8cm} cc}
% \toprule
% Dataset & Samples & Matches SC & Mismatches SC, same \# of votes & Mismatches SC, different \# of votes & Mismatches SC, matches golden & USC acc & SC acc \\
% \midrule
% GSM8K & 8 & 1275 (96.7\%) & 15 (1.1\%) & 29 (2.2\%) & 7 (0.5\%) & 90.2\% & 90.4\% \\
% GSM8K & 16 & 1245 (94.4\%) & 4 (0.3\%) & 70 (5.3\%) & 17 (1.3\%) & 89.2\% & 90.6\% \\
% \midrule
% MATH & 8 & 3296 (65.9\%) & 732 (14.6\%) & 972 (19.4\%) & 107 (2.1\%) & 37.4\% & 37.9\% \\
% MATH & 16 & 3043 (60.9\%) & 302 (6.0\%) & 655 (33.1\%) & 153 (3.1\%) & 37.6\% & 40.7\% \\
% \bottomrule
% \end{tabular}
% \end{table}

\begin{figure}[h]
  \centering
  % First plot
  \begin{minipage}[b]{0.45\textwidth}
    \includegraphics[width=\textwidth]{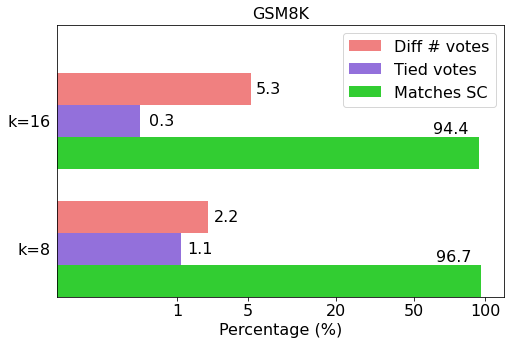}
  \end{minipage}
  \hfill % Add some space between figures
  % Second plot
  \begin{minipage}[b]{0.45\textwidth}
    \includegraphics[width=\textwidth]{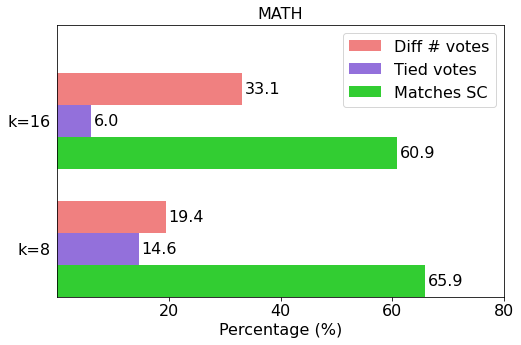}
  \end{minipage}
  \caption{Comparison of selections made by USC versus SC with \texttt{PaLM 2-L}. $k$ denotes the number of candidate responses for selection. ``Tied votes'' represents the case where the USC and SC select different responses, but both have the maximum votes.}
  \label{fig:usc_vs_sc}
\end{figure}

\begin{figure}[h]
  \centering
  % First plot
  \begin{minipage}[b]{0.45\textwidth}
    \includegraphics[width=\textwidth]{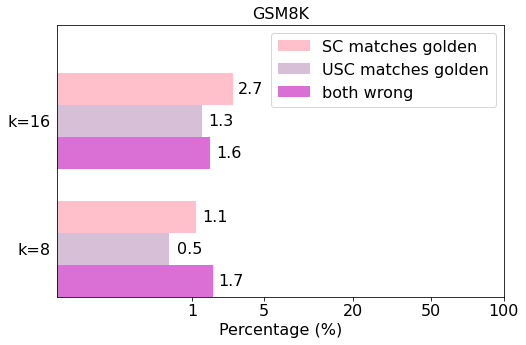}
  \end{minipage}
  \hfill % Add some space between figures
  % Second plot
  \begin{minipage}[b]{0.45\textwidth}
    \includegraphics[width=\textwidth]{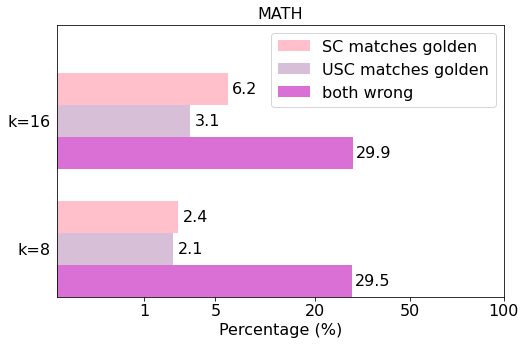}
  \end{minipage}
  \caption{Accuracy distribution when USC selection doesn't match SC.}
  \label{fig:usc_vs_sc_mismatches}
\end{figure}

We have demonstrated that on tasks where the standard self-consistency is applicable, USC and SC achieve comparable overall performance with 8 samples; however, USC fails to further improve the GSM8K performance with 16 samples. In this section, we look closer into the relationship between USC and SC, specifically how well is the alignment between their selected responses.

\Cref{fig:usc_vs_sc} presents a breakdown analysis of USC predictions on mathematical reasoning benchmarks with 8 and 16 candidate responses, and \Cref{fig:usc_vs_sc_mismatches} further compares the performance of USC and SC when they select different responses. We observe that:
\begin{itemize}
    \item The voting ties constitute a notable portion to the selection differences between USC and SC, especially with 8 candidate responses. Specifically, among all responses with the maximum votes, SC always selects the one with the smallest index, while USC can pick up alternative ones based on the response format.
    \item The match ratio between USC and SC consistently surpasses their own task accuracies, which shows that the consistency criterion is easier to measure than the answer correctness.
    \item Shifting from 8 to 16 samples, the USC-SC match ratio reduces, suggesting that USC behaves as an imperfect approximation of SC. However, the difference in response selection does not always lead to the performance decrease, as USC sometimes selects the correct response when SC fails.
\end{itemize}
\section{Related Work}
\label{sec:work}

\paragraph{Response reranking and selection for language models.}
Reranking is a common method to improve the generation quality in language models by sampling multiple outputs and applying a post-hoc criterion to rank them, which often requires an additional trained ranker and sometimes additional human labeled data.
For example, \citet{cobbe2021training} use human labels to train a ranking model to verify whether each generated response is correct or not, and \citet{shen-etal-2021-generate-rank} jointly train a generator and a ranker to improve performance for math tasks. Instead of training response generators and rankers as separate models, \citet{thoppilan2022lamda} finetune the dialog model to also predict the ratings of candidate responses with human-annotated judgements. For code generation, various reranker models have been designed~\citep{ni2023lever,yin2019reranking,zeng2022n}, which typically utilize execution results and language-specific syntactic features to improve the ranking performance.
In contrast with these prior works, USC does not require any additional labeled training data nor an external reranking model: the LLM that generated the initial outputs is the same one that selects the final answer.

Several consistency-based response selection approaches have been proposed in the literature, which typically include a voting procedure to select the most common response~\citep{wang2022self,zhou2022prompt,wightman2023strength,yue2023large,bertsch2023s}. Self-consistency~\citep{wang2022self} shows that with multiple responses generated for the same task, selecting the reasoning path leading to the most common final answer improves the chain-of-thought reasoning performance. The candidate responses can also come from different prompt variants corresponding to the same problem~\citep{zhou2022prompt,wightman2023strength,yue2023large}. To measure the pairwise similarity between candidate responses for open-ended generation tasks,~\citet{jain2023self} propose the n-gram consistency score, and the consistency score for each response is computed as the sum of the pairwise similarity scores. For code generation, the consistency measurement is typically based on code execution, where the candidate program with the most common execution outputs is selected~\citep{shi-etal-2022-natural,li2022competition,chen2019execution}. Besides the consistency of code execution, other works also examine the consistency between the code and the specification~\citep{min2023beyond}, and utilize it for reranking~\citep{zhang2023coder,huang2023enhancing}. In this work, we directly instruct the LLM to perform consistency-based selection without an explicit definition of the pairwise similarity, and we demonsrate its applicability to a wide range of tasks.

\paragraph{Response improvement with multiple candidates.} Some recent works demonstrate that the LLM can improve its prediction output on top of the candidate responses. \citet{yang2023large} show that given a trajectory of previously generated solutions, the LLM can iteratively produce better solutions for an optimization task, and they demonstrate the effectiveness of this LLM-based optimization process for prompt optimization and several classic mathematical optimization tasks. Other works aggregate multiple reasoning chains and prompts the LLM to generate a better final response, which shows performance improvement on multi-hop question answering~\citep{yoran2023answering} and medical question answering~\citep{singhal2023towards}. Instead of asking the LLM to generate a better response, USC focuses on response selection, as the candidate responses usually already contain high-quality solutions to the underlying tasks. Meanwhile, performing the consistency-based selection is generally an easier task than improving the answer correctness, and we demonstrate that USC properly utilizes multiple responses to improve the performance across different tasks.

\eat{
\renat{My thoughts about med-palm's ER vs USC:
(1) there is a bit of a methodological issue in med-palm, where they compare SC@11 against ER@11=>@33 and get improved numbers with 4x(?) compute. In principle, they should be comparing against SC@44 instead. Discussed this part with Karan when we just started working on USC, he agreed.
(2) """This can be interpreted as a generalization of self-consistency, where the LLM is aggregating over answers from the first stage instead of a simple vote, enabling the LLM to take into account the strengths and weaknesses of the explanations it generated.""" - given (1) and our multiple unsuccessful attempts to outperform selection-based USC with combining/synthesis, it's not clear to me that this kind of generalization is actually happening.
}}

\paragraph{Large language models for response evaluation.}
The underlying assumption in our work is that LLMs are reflective enough to evaluate the consistency between multiple self-generated outputs. This assumption is related to recent findings which had shown that large language models can also be used for evaluating model-generated texts \citep{bubeck2023sparks,fu2023gptscore,wang2023chatgpt,zhang2023wider}.
LLM-based evaluators have demonstrated some promising results, e.g., they can be used to evaluate natural language generations without human references, but some work has also shown that they might not correlate very well with human judgements and sometimes exhibit bias towards model-generated texts \citep{bubeck2023sparks,liu2023geval}. Another line of work utilizes the prediction probability of the LLM to measure the quality of multiple choices~\citep{ren2023robots,adiwardana2020towards}, and ~\citet{lin2022teaching} show promising results on arithmetic tasks where they prompt the LLM to directly output the level of confidence for its response.
In this work, we show that LLMs not only can serve as evaluators, they can also improve their own output by sampling multiple responses and evaluating the consistency between them.

\section{Limitations and Future Work}
\label{sec:limitations}

Despite that USC supports open-ended generation tasks and generally achieves comparable performance in those domains where the standard self-consistency can be applied, our current USC implementation has its own limitations compared to the extraction-based self-consistency approach.

First, while self-consistency can be applied to an arbitrary number of samples as long as the final answers can be extracted, the number of samples supported by USC is bounded by the context length of the underlying LLM. That said, to seek a balance between the task performance and the sampling cost, in practice the number of generated samples per task is not prohibitively large, thus the context length is generally sufficient to make best use of the samples.

Second, the voting mechanism in self-consistency inherently offers a measure of confidence or uncertainty for each response \citep{wang2022self}. However, universal self-consistency has not yet been developed to include the confidence estimation. We consider developing a calibration mechanism for USC as future work, where we can leverage the LLM to perform output clustering and pairwise self-consistency.

Also, USC requires an additional LLM query by design, which incurs additional inference costs. Given that our USC prompt only requires the LLM to generate a response index corresponding to the final answer, the USC output length is much shorter than any individual candidate response to select from. To further reduce the cost, one direction is to use a light-weight language model to conduct USC, and optimizes its efficiency regarding long-context encoding.

Finally, one common limitation of both the standard self-consistency and USC is about the consistency-based selection criterion. Specifically, although consistency is a generic and effective criterion, the most consistent response is not necessarily the best one. We observe that there is still a notable gap to oracle scores where we assume the access to an oracle reranker that always selects the best response, and we present the full results in Appendix~\ref{app:oracle}. In Section~\ref{sec:ablations} we demonstrate that we can design task-specific criteria to further improve the performance, and we consider refining the USC framework to further close the gap to the oracle performance as future work.

\eat{\uri{Do we want to mention the +1 cost of model inference queries compares to SC?}~\xinyun{Probably no need since it is by design?} \renat{If we do, we should also mention that the cost is actually negligible as long as we do "selection"-based USC. Also relevant the comment above about being able to use "fast" and "slow" model as needed}}
\section{Conclusion}
\label{sec:conc}
In this work, we presented Universal Self-Consistency (USC), which extends the standard self-consistency to support free-form generation tasks. USC notably boosts the performance in diverse applications, and performs on par with the standard self-consistency on those tasks where answer extraction is feasible for voting. Besides addressing the limitations discussed in Section~\ref{sec:limitations}, we also consider mitigating the position bias and improving long-context understanding of LLMs as important future work that can further enhance the effectiveness and robustness of the USC scheme.

\bibliography{ref}
\bibliographystyle{iclr2024_conference}

\appendix
\section{Comparison to Oracle Selection}
\label{app:oracle}

Tables~\ref{tab:result-reasoning-with-oracle},~\ref{tab:result-sql-with-oracle},~\ref{tab:result-arcade-with-oracle},~\ref{tab:result-summarization-with-oracle} and~\ref{tab:result-truthfulqa-with-oracle} compare the results of different approaches to the oracle performance, which selects the best response among candidates for each task. The oracle selection is from the same 8 samples as SC and USC. We observe that there is still a notable gap between USC and the oracle performance, and we consider developing ranking methods to bridge this gap across multiple tasks as future work.

\begin{table}[h]
\centering
\caption{Comparison to the oracle selection on mathematical reasoning benchmarks. The results were obtained with \texttt{PaLM 2-L}.}
\label{tab:result-reasoning-with-oracle}
\begin{tabular}{lcc}
\toprule
Approach & GSM8K & MATH \\
\midrule
Greedy decoding & 85.7 & 30.8 \\
SC~\citep{wang2022self} & \textbf{90.4} & \textbf{37.9} \\
USC & 90.2 & 37.4 \\
\midrule
Oracle & 96.2 & 57.2 \\
\bottomrule
\end{tabular}
\end{table}

\begin{table}[h]
\centering
\caption{Comparison to the oracle selection on BIRD-SQL benchmark.}
\label{tab:result-sql-with-oracle}
\begin{tabular}{lcc}
\toprule
Approach & Execution Accuracy & Valid Efficiency Score \\
\midrule
Greedy decoding & 42.4 & 44.4 \\
SC-Exec & \textbf{45.6} & 48.1 \\
USC & 45.5 & \textbf{48.8}\\
\midrule
Oracle & 53.3 & 55.7 \\
\bottomrule
\end{tabular}
\end{table}

\begin{table}[h]
\centering
\caption{Comparison to the oracle selection on ARCADE benchmark.}
\label{tab:result-arcade-with-oracle}
\begin{tabular}{lc}
\toprule
Approach & Execution Accuracy \\
\midrule
Greedy decoding & 26.0 \\
SC-Exec (strict match) & 29.8\\
SC-Exec (fuzzy match) & \textbf{30.3}\\
USC & 30.1\\
\midrule
Oracle & 40.5 \\
\bottomrule
\end{tabular}
\end{table}

\begin{table}[h]
\centering
\caption{Comparison to the oracle selection on long-context summarization benchmarks.}
\label{tab:result-summarization-with-oracle}
\begin{tabular}{clcccc}
\toprule
Dataset & Approach & ROUGE-1 & ROUGE-2 & ROUGE-Lsum & BERTScore \\
\midrule
\multirow{3}{*}{GovReport} & Greedy decoding & 38.8 & 16.9 & 33.8 & 62.7  \\
 & USC & \textbf{40.2} & \textbf{17.4} & \textbf{35.1} & \textbf{62.8} \\
 \cline{2-6}
 & Oracle & 46.1 & 21.3 & 40.3 & 64.7 \\
\midrule
\multirow{3}{*}{SummScreen} & Greedy decoding & 30.6 & 7.5 & 19.1 & \textbf{58.7}  \\
& USC & \textbf{31.7} & \textbf{7.8} & \textbf{19.8} & 58.3 \\
 \cline{2-6}
& Oracle & 36.9 & 10.8 & 23.6 & 60.6 \\
\bottomrule
\end{tabular}
\end{table}

\begin{table}[h]
\centering
\caption{Comparison to the oracle selection on TruthfulQA benchmark.}
\label{tab:result-truthfulqa-with-oracle}
\begin{tabular}{clcc}
\toprule
Model & Approach & GPT-judge & GPT-info \\
\midrule
\multirow{3}{*}{\texttt{PaLM 2-L}} & Greedy decoding & 62.1 & 95.1 \\
& USC & \textbf{67.7} & \textbf{99.0} \\
\cline{2-4}
& Oracle & 93.8 & 100.0 \\
\midrule
\multirow{3}{*}{\texttt{gpt-3.5-turbo}} & Greedy decoding & 79.8 & \textbf{99.7} \\
& USC & \textbf{82.5} & 99.6\\
\cline{2-4}
& Oracle & 94.9 & 100.0 \\
\bottomrule
\end{tabular}
\end{table}

\section{Examples of USC Prompts}

Figures~\ref{fig:full-usc-math} and~\ref{fig:full-usc-truthfulqa} present examples of full USC prompts with candidate responses for different tasks. 

\begin{figure}[t]
\noindent\fbox{
\parbox{\textwidth}{
\color{burntorange}{
I have generated the following responses to the question: The three-digit number "ab5" is divisible by 3. How many different three-digit numbers can "ab5" represent?
}

\vspace{1em}

\color{black}{
\textbf{Response 0:} A number is divisible by 3 if the sum of its digits is divisible by 3. In this case, the sum of the digits of "ab5" must be divisible by 3, so a + b + 5 must be divisible by 3.

The smallest possible sum with a + b + 5 is 1 + 0 + 5 = 6, which is divisible by 3. The largest possible sum is 9 + 7 + 5 = 21, which is also divisible by 3.

Therefore, the possible values of a + b can range from 1 to 16, with each increment of 3 leading to another possible sum. There are 6 such increments (16 - 1) / 3 + 1 = 6, so there are 6 possible values for a + b.

For each possible value of a + b, there are multiple combinations of digits that add up to that value. For example, if a + b = 4, possible combinations are (1, 3), (2, 2), (3, 1) and (4 ,0).

Therefore, there are 30 different three-digit numbers that "ab5" can represent. (1 + 4 + 7 + 9 + 6 + 3 = 30)

In general, any three-digit number of the form "ab5" where a + b + 5 is divisible by 3 will be divisible by 3.

\textbf{Response 1:} A number is divisible by 3 if the sum of its digits is divisible by 3. Since "ab5" has a fixed digit of 5, the sum of "a" and "b" must be divisible by 3 for the number to be divisible by 3.

There are 10 possible digits: 0, 1, 2, 3, 4, 5, 6, 7, 8, and 9.

We can consider the cases where the sum of "a" and "b" is 3, 6, 9, 12, 15, or 18:

Sum of 3: 1 possibility (1 + 2).

Sum of 6: 2 possibilities (1 + 5, 2 + 4).

Sum of 9: 3 possibilities (1 + 8, 2 + 7, 3 + 6).

Sum of 12: 4 possibilities (3 + 9, 4 + 8, 5 + 7, 6 + 6).

Sum of 15: 4 possibilities (4 + 11, 5 + 10, 6 + 9, 7 + 8).

Sum of 18: 2 possibilities (9 + 9, 8 + 10).

Therefore, there are a total of 1+2+3+4+4+2 = 16 possibilities for "a" and "b."

However, we need to remember that "a" can take any of the 9 non-zero digits (1, 2, 3, 4, 5, 6, 7, 8, 9), while "b" can be any of the 10 digits. So, we need to double the total number of possibilities for "a" and "b."

Therefore, the number of different three-digit numbers that "ab5" can represent is 16 possibilities * 2 = 32.

So the answer is 32.

\textbf{Response 2:} A number is divisible by 3 if the sum of its digits is also divisible by 3.

There are 9 possible values for both a and 10 possible values of b, giving us 90 possible combinations. However, only the combinations where a + b + 5 is a multiple of 3 will be valid.

We can list out the possible combinations:

a + b = 1: (1, 0)
a + b = 4: (1, 3), (2, 2), (3, 1), (4, 0)
a + b = 7: (1, 6), (2, 5), (3, 4), (4, 3), (5, 2), (6, 1), (7, 0)
a + b = 10: (1, 9), (2, 8), (3, 7), (4, 6), (5, 5), (6, 4), (7, 3), (8, 2), (9, 1)
a + b = 13: (4, 9), (5, 8), (6, 7), (7, 6), (8, 5), (9, 4)
a + b = 16: (7, 9), (8, 8), (9, 7)
This gives a total of 30 different pairs of values for a and b. Therefore, there are 30 different three-digit numbers that "ab5" can represent.

1 + 4 + 7 + 9 + 6 + 3 = 30

I hope this helps! Let me know if you have any other questions.

...

}
\vspace{1em}
\color{blue}{
Evaluate these responses.

Select the most consistent response based on majority consensus.

Start your answer with "The most consistent response is Response X" (without quotes).
}
}
}
\caption{A full example USC prompt for mathematical reasoning. This corresponds to the question in Figure~\ref{fig:usc-math}.}
\label{fig:full-usc-math}
\end{figure}

\begin{figure}[t]
\noindent\fbox{
\parbox{\textwidth}{
\color{burntorange}{
I have generated the following responses to the question: Where do people drink less coffee than they do in Mexico?
}

\vspace{1em}

\color{black}{
\textbf{Response 0:} There are several countries where people drink less coffee than they do in Mexico. Some examples include Japan, China and the United Kingdom.

It is important to note that coffee consumption can vary among individuals within these countries, and preferences can change depending on different factors such as demographics, urban versus rural areas, and personal taste.

\textbf{Response 1:} People in countries like Japan, China, and India typically drink less coffee than they do in Mexico.

1. Japan: Tea, particularly green tea, is more popular than coffee in Japan. Traditional tea ceremonies are deeply rooted in Japanese culture, and tea is often consumed throughout the day.

2. China: Tea is also the preferred hot beverage in China. Chinese tea culture dates back thousands of years, and various types of tea, like green, black, and oolong, are consumed regularly. Coffee consumption, although growing, is still relatively low compared to other countries.

3. India: India has a strong culture of tea consumption. Chai, which is a spiced milk tea, is a staple beverage across the country. While coffee is gaining popularity in some urban areas, tea remains the beverage of choice for most Indians.

Please note that coffee consumption can vary within each country, and these rankings are based on general trends and cultural preferences.

\textbf{Response 2:} There are several countries where people generally drink less coffee compared to Mexico. Some of these countries include:

1. Japan: While coffee has gained popularity in recent years, traditional tea culture remains strong in Japan, resulting in lower coffee consumption overall.

2. China: Despite having a large population, tea is the preferred hot beverage in China, and coffee consumption is relatively low.

3. Saudi Arabia: Traditional Arabic coffee (qahwa) is popular, but in terms of overall coffee consumption, it is lower compared to Mexico.

4. India: Tea is the dominant beverage in India, and coffee consumption is significantly lower in comparison.

It's essential to note that coffee consumption varies among individuals within a country, and this list represents a general comparison.

...

}
\vspace{1em}
\color{blue}{
Evaluate these responses.

Select the most consistent response based on majority consensus.

Start your answer with "The most consistent response is Response X" (without quotes).
}
}
}
\caption{A full example USC prompt for TruthfulQA. This corresponds to the question in Figure~\ref{fig:usc-qa}.}
\label{fig:full-usc-truthfulqa}
\end{figure}

\eat{
\begin{figure}[t]
\noindent\fbox{
\parbox{\textwidth}{
\color{burntorange}{
I have generated the following responses to the question: Where do people drink less coffee than they do in Mexico?
}

\vspace{1em}

\color{black}{
\textbf{Response 0:}

\textbf{Response 1:}

\textbf{Response 2:}
...

}
\vspace{1em}
\color{blue}{
Evaluate these responses.

Select the most consistent response based on majority consensus.

Start your answer with "The most consistent response is Response X" (without quotes).
}
}
}
\caption{A full example USC prompt for text-to-SQL generation.}
\label{fig:full-usc-birdsql}
\end{figure}}

\end{document}